\documentclass{article}
\usepackage[margin=1in]{geometry}

\usepackage{graphicx}
\usepackage{booktabs}
\usepackage{amsmath}
\usepackage{amssymb}
\usepackage{amsfonts}
\usepackage{multirow}
\usepackage{verbatim}
\usepackage{caption}
\usepackage{longtable}
\usepackage{supertabular}
\usepackage{float}

\usepackage{enumitem}
\usepackage{tablefootnote}
\usepackage[round,semicolon]{natbib}
\usepackage{xcolor}
\usepackage{xspace}
\usepackage{textcomp}
\usepackage{makecell}
\usepackage{multirow}
\usepackage{lscape} 
\usepackage{siunitx}

\usepackage{amssymb}%
\usepackage{pifont}%
\usepackage{scrextend}

\usepackage{array}
\usepackage{tgpagella}
\usepackage{latexsym}
\usepackage[T1]{fontenc}
\usepackage[utf8]{inputenc}
\usepackage{microtype}
\definecolor{mydarkblue}{rgb}{0,0.08,0.45}
\usepackage[colorlinks,citecolor=mydarkblue,urlcolor=mydarkblue,linkcolor=mydarkblue]{hyperref}
\usepackage{url}         
\usepackage{nicefrac}       
\usepackage{changepage}
\usepackage{xargs}          
\usepackage{wrapfig,lipsum,booktabs}
\usepackage{longtable}
\usepackage{endnotes}

\usepackage{pgfplots}
\usetikzlibrary{pgfplots.groupplots}
\pgfplotsset{compat=1.3}
\usepackage{tikz}
\usetikzlibrary{patterns}

\usepackage[most]{tcolorbox}

\usepackage[capitalize,noabbrev]{cleveref}
\crefname{section}{Section}{\S\S}
\Crefname{section}{Section}{\S\S}
\crefname{table}{Table}{Tables}
\crefname{figure}{Figure}{Figures}
\crefname{algorithm}{Algorithm}{}
\crefname{equation}{eq.}{}
\crefname{appendix}{Appendix}{}
\crefformat{section}{Section #2#1#3}
\usepackage{multicol}
\usepackage{subcaption}

\DeclareMathOperator{\diag}{diag}

\usepackage{minitoc}

\usepackage{microtype}
\usepackage{graphicx}
\usepackage{booktabs} %
\usepackage{hyperref}
\usepackage{siunitx}
\usepackage{amsmath}
\usepackage{amssymb}
\usepackage{mathtools}
\usepackage{amsthm}
\usepackage{fontawesome5}
\usepackage{longtable}
\usepackage{multirow}
\usepackage{caption}
\usepackage{xfrac}
\usepackage{setspace}

\usepackage{wrapfig}
\usepackage{tabulary}
\usepackage{tabularx}
\usepackage[export]{adjustbox}

\usepackage[capitalize,noabbrev]{cleveref}

\newcommand{\method}{\text{FHLR}}

\title{
\textbf{Learning under Label Noise through Few-Shot Human-in-the-Loop Refinement}
}
  
\author{
\normalsize{}
\textbf{Aaqib Saeed$^1\footnote{Correspondence: a.saeed@tue.nl}$, 
 Dimitris Spathis$^{2,3}$, Jungwoo Oh$^4$, Edward Choi$^4$, Ali Etemad$^5$}\\
\normalsize{}
$^1$Eindhoven University of Technology, $^2$Nokia Bell Labs, $^3$University of Cambridge, $^4$KAIST, $^5$Queen's University
}

\date{}

\begin{document}
\maketitle

\begin{abstract}
Wearable technologies enable continuous monitoring of various health metrics, such as physical activity, heart rate, sleep, and stress levels. A key challenge with wearable data is obtaining quality labels. Unlike modalities like video where the videos themselves can be effectively used to label objects or events, wearable data do not contain obvious cues about the physical manifestation of the users and usually require rich metadata. As a result, label noise can become an increasingly thorny issue when labeling such data. In this paper, we propose a novel solution to address noisy label learning, entitled \textit{Few-Shot Human-in-the-Loop Refinement} (FHLR). Our method initially learns a seed model using weak labels. Next, it fine-tunes the seed model using a handful of expert corrections. Finally, it achieves better generalizability and robustness by merging the seed and fine-tuned models via weighted parameter averaging. We evaluate our approach on four challenging tasks and datasets, and compare it against eight competitive baselines designed to deal with noisy labels. We show that FHLR achieves significantly better performance when learning from noisy labels and achieves state-of-the-art by a large margin, with up to $19\%$ accuracy improvement under symmetric and asymmetric noise. Notably, we find that FHLR is particularly robust to increased label noise, unlike prior works that suffer from severe performance degradation. Our work not only achieves better generalization in high-stakes health sensing benchmarks but also sheds light on how noise affects commonly-used models. 
\end{abstract}

\section{Introduction}
The increasing adoption of wearable technology has enabled continuous monitoring of various health metrics, such as physical activity, heart rate, sleep, and stress levels. This has spurred interest in gleaning insights into health and wellness from the data collected by these ubiquitous devices, for instance by detecting potential complications and promoting healthy behaviors. Beyond personal use, data coming from wearables also have promising medical applications. Physicians can monitor the health of patients remotely, especially those with chronic conditions, and track their progress over time. This is particularly useful for detecting changes that require prompt medical attention. Moreover, the continuous physiological data from wearables along with other devices can help doctors make more accurate diagnoses and develop personalized treatment plans. 

The abundance of data generated from wearable sensors has paved the way for developing deep learning models to tap into these insights. However, deep models rely on large volumes of high-quality, clean, and labeled data, which can be difficult to obtain in the context of wearable signals. Data labels are not always accurate due to factors like users’ subjective interpretations, lack of domain expertise, and annotation cost. Inconsistent labels can undermine the generalizability of deep learning models, especially for health monitoring where misdiagnosis can have grave consequences. Therefore, developing techniques to mitigate the effects of noisy labels is crucial to fully realize the potential of deep learning for wearable time-series. 

While there has been significant research on mitigating label noise in deep learning (see Section~\ref{sec:related_work}) in the context of language and vision, to the best of our knowledge, there has been no rigorous attempt at dealing with label corruption in the context of deep networks for wearable sensor data. To this end, we aim to investigate the impact of label noise and provide an effective way to mitigate its impact on training deep neural networks for sensory data, such as signals obtained from inertial measurement units (IMUs), electroencephalography (EEG), and electrocardiography (ECG). 

\begin{figure*}[t]
    \centering
    \includegraphics[width=\textwidth]{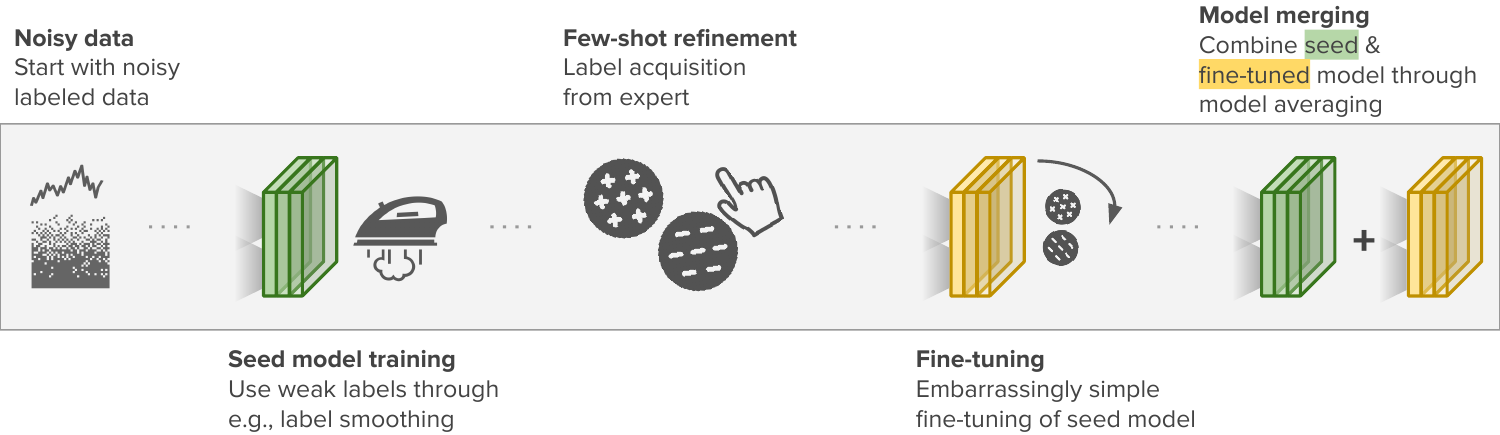}
    \caption{\textbf{Illustration of the proposed noisy label learning framework.} Overview of~\method, a simple yet effective method for dealing with noisy labels by pre-training a model with weak labels, fine-tuning with expert annotations, and performing weight averaging to come up with the final model.}
    \label{fig:overview}
\end{figure*}

In this paper, we propose a novel technique to tackle the issue of noisy labels, named \textit{Few-shot Human-in-the-Loop Refinement} (\method). Our training scenario consists of three main stages (see Figure~\ref{fig:overview} for an overview). In the initial phase, our approach learns a seed model with weak labels, where labels are generated by smoothing existing noisy ones in order to obtain a reliable initial model that is less prone to overfitting to noisy annotations. In the second phase, we leverage a small set of clean labels acquired from human experts to fine-tune the initial model. As a last step, we apply model merging (i.e., weighted averaging of learned model parameters) to create a more accurate and robust model with better generalization. We evaluate our method across four tasks and datasets, comparing against eight baselines, and show that \method~provides significantly better generalization in learning from noisy labeled data than prior techniques.

\method~aims to enhance the performance of deep models in the context of wearable time-series with noisy labels, while offering several advantages over traditional methods. First, it does not make any assumptions about the distribution of label noise, which makes it applicable to various real-world noise profiles and different modalities. Second, incorporating human expertise into the annotation process ensures that the labels are grounded in domain knowledge, and yet, our `few shot' approach does not require extensive involvement from the experts and also enable the mitigation of annotation noise present in real-world datasets. Third, building upon the success of weight averaging of fine-tuned neural networks~\citep{wortsman2022model} and learning from weak supervision~\citep{lukasik2020does}, our approach offers a powerful way for learning from noisy labeled data. To the best of our knowledge, model averaging has not been explored earlier in learning under label noise. Finally, our method does not rely on any supplementary neural networks or modified loss functions, incurs no additional costs during inference, and enables efficient training of a robust model using a minimal number of clean examples that can be practically available in practice. 
In this work, we make the following contributions:
\begin{itemize}
    \item \textbf{High-impact domain}. We study for the first time the effect of label noise on learning models from wearable sensor data in the context of health and well-being tasks, such as sleep-stage scoring. 
    \item \textbf{Novel effective method.} We propose a highly effective method (\method) for addressing noisy labels through few-shot human-in-the-loop refinement that outperforms several prior techniques. 
    \item \textbf{Particularly robust to high noise levels}. We empirically demonstrate that our approach yields models with high generalizability and provides robustness against low to high levels of noise in the label space.
    \item \textbf{Strong results in an array of competitive benchmarks}. We show that our embarrassingly simple averaging of seed and fine-tuned models exhibit better performance than individual counterparts and on-par with computationally expensive methods, such as model ensembles.   
\end{itemize}

\section{Related Work}
\label{sec:related_work}
Over the years, there has been significant interest in the study of learning deep models from noisy labels within the scope of data-centric and robust deep learning. In order to minimize the effects of label noise, recent research has explored various strategies. One approach is to use regularization techniques such as dropout~\citep{arpit2017closer} which can help prevent overfitting and improve generalization. Label cleaning and correction techniques have also been proposed~\citep{reed2014training, goldberger2017training, li2017learning, veit2017learning, song2019selfie}, where additional steps are performed to identify and correct mislabeled instances in the training data.

Another strategy is instance re-weighting, where the contribution of each training instance to the learning objective is adjusted based on its estimated reliability or confidence. Mentornet~\citep{jiang2018mentornet}, Co-teaching~\citep{ren2018learning}, and Meta-weight-net~\citep{shu2019meta} are examples of methods that assign different weights to instances based on their predicted probabilities or distances to decision boundaries. Cross-validation has also been used to tackle label noise~\citep{northcutt2021confident}, where the training data is divided into multiple subsets and models are trained and evaluated on different subsets to identify samples with incorrect labels. Other approaches include meta learning~\citep{zheng2021meta}, self-learning~\citep{han2019deep}, gradient clipping~\citep{menon2020can}, and data augmentation~\citep{zhangmixup,cheng2020advaug,liang2020simaug,jiang2020beyond}, all of which have been investigated to mitigate the effects of label noise.

In the context of the vision domain, label smoothing~\citep{lukasik2020does} has been studied to tackle label noise. However, its feasibility for modalities like sequential data remains uncertain, and this is the gap our method aims to fill. Additionally, while the cross-validation approach~\citep{northcutt2021confident} has been successful in identifying samples with incorrect labels, it comes with a significant training cost which may result in discarding a large number of valuable training examples. Further, although model merging has been explored in the context of ensemble learning, it has not been widely used to tackle label noise.

To the best of our knowledge, no prior works have specifically studied wearable time-series representation learning under label noise, additionally assessing both symmetric (where mislabeling occurs randomly) and asymmetric (where mislabeling occurs systematically or in a biased manner) noise. Further, wearable sensing lacks reliable crowdsourcing unlike vision, where, cheap image labels can be acquired via Mechanical Turk. Our work addresses this shortcoming by proposing a highly effective yet simplistic approach. Our method is complementary to existing approaches, as it does not modify the primary objective function and can be used as a plug-in to achieve high performance in learning from noisy labels.

\section{Method}
\subsection{Preliminaries}
\textbf{Label Noise.} Label noise refers to the misalignment between the ground truth label $y^{*}$ and the observed label $y$ in a given dataset. In the context of a $\mathcal{C}$-way classification problem, label noise can be modeled as a class-conditional label flipping process $h\left(\cdot\right)$, where every label in class $j\in\mathcal{C}$ may be independently mislabeled as class $i\in\mathcal{C}$ with probability $p\left(y=i{\mid}y^{*}=j\right)$, denoted by $p\left(y{\mid}y^{*}\right)$. Here, we assume that the instances of label noise are data-independent, meaning that $p\left(y{\mid}y^{*}, x\right) = p\left(y{\mid}y^{*}\right)$, in line with previous work~\citep{northcutt2021confident}. The label noise function $h\left(y^{*}, \mathcal{C}\right)$ allows for the definition of a $\mathcal{C}\times\mathcal{C}$ noise distribution matrix, denoted by $\mathcal{Q}_{y{\mid}y^{*}}$, where each column represents the probability distribution for an input instance with ground truth label $y^{*} = i$ to be assigned to label $j$. This matrix captures the inherent uncertainty in the labeling process and can be used to study the effects of label noise on the performance of learning algorithms. Based on these particulars, we can describe label noise through the following statistical parameters: noise level $n_{l}$ and noise sparsity $n_{s}$. 

The label noise level ($n_l$) quantifies the extent of inaccurate labels present in a given data corpus. Intuitively, a noise level of zero corresponds to a ``pristine'' dataset, in which all observed labels correspond to their true labels, while a noise level of one would represent a completely erroneous dataset. Formally, it is defined as one minus the diagonal sum of the conditional probability matrix $\mathcal{Q}_{y{\mid}y^{*}}$, denoted as $n_{l} = 1 - \diag \left( \mathcal{Q}_{y{\mid}y^{*}} \right)$. Similarly, the noise sparsity ($n_s$) quantifies the structure of label noise present in a dataset. It is defined as the fraction of zeros in the off-diagonals of the noise distribution matrix $\mathcal{Q}_{y{\mid}y^{*}}$~\citep{northcutt2021confident}. Therefore, a high noise sparsity values indicate a non-uniformity of label noise, which is common in most real-world datasets. Specifically, zero noise sparsity corresponds to a random noise, confounding instances across classes. The special case of `class-flipping' occurs at $n_s=1$, confusing instance pairs.

\subsection{Problem Setup and Formulation}
We consider a general $\mathcal{C}$-way classification as a supervised learning task that aims to learn a function mapping an input instance $\mathbf{x}$ to a corresponding ground truth label $\mathbf{y}_i^* \in \{1, \cdots, \mathcal{C}\}$. The input space is denoted by $\mathcal{X}=\mathbb{R}^d$ and the output space by $\mathcal{Y}=\{1,2,..., \mathcal{C}\}$. The ground truth can also be formulated as a one-hot encoded vector $\mathbf{y}_i^* \in \{0,1\}^\mathcal{C}$, where $\mathcal{C}$ is the number of classes. This indicates that the $\mathbf{y}_i^*$ is a binary vector of length $\mathcal{C}$, where only one element is 1 and the rest are 0s, representing the true class of the sample $i$ among $\mathcal{C}$ possible classes. A learner is given access to a set of training data $\mathcal{D}={(\mathbf{x}_i,\mathbf{y}_i)}_{i=1}^N$ drawn from an unknown joint data distribution $\mathcal{P}$ defined on $\mathcal{X}\times \mathcal{Y}$. A neural network $f(\mathbf{x};\theta): \mathcal{X} \to \mathbb{R}^\mathcal{C}$ minimizes the empirical risk:
\begin{align*}
    R_\mathcal{L}(f)=\mathbb{E}_{D}(\mathcal{L}_\text{CE}(f(\mathbf{x};\theta),\mathbf{y})),
\end{align*}
\noindent where $\theta$ are the parameters of the network, and $\mathcal{L}$ is a loss function.

Specifically, $f_y(\mathbf{x})$ denotes the $y$-th element of $f(\mathbf{x})$ corresponding to the ground-truth label $\mathbf{y}$. 
When $n_l \neq 0$, the neural network $f(\mathbf{x};\theta)$ is trained on labels $\mathbf{y}$, instead of the actual ground-truth labels $\mathbf{y}^*$. 
As a result, our objective is to design a mitigation strategy that can either correct noisy labels or provide a way to reduce their impact on the learning process. Moreover, the ideal solution should maintain consistent performance for different values of $0 \le n_l \le 1$ to eliminate the need for any prior assumptions regarding the level and distribution of noise.

\subsection{Few-shot Human-in-the-Loop Refinement}
We introduce \textit{Few-shot Human-in-the-Loop Refinement} (\method), a highly effective approach to training deep neural networks for sensory (time-series) data with noisy labels. \method~ enables an efficient way to incorporate a few expert labels for fine-tuning a seed model and apply weight averaging to merge models in order to improve generalization. In this section, we describe an end-to-end pipeline and provide a high-level overview of the approach in Figure~\ref{fig:overview}. %

\textbf{Seed Training with Weak Labels.} We begin with bootstrapping a deep model $f_{\mathcal{B}}(\cdot)$ with noisy labeled data without discarding any instances. Rather than using strongly labeled data (which may have noise) directly at this stage, we generate `weak labels' through label smoothing (LS)~\citep{szegedy2016rethinking}, which produces softened one-hot encoded vectors representing semantic information. These weak labels aid in the initial training of neural networks, allowing them to learn representations even when all labels are not high quality. Label smoothing creates weak labels by replacing hard labels (a single index indicating a class) with softened labels (a vector of weights that sum to 1, indicating the degree of class membership). 

Formally, for a dataset $\mathbf{X} = \{\mathbf{x}_1, \mathbf{x}_2, \ldots, \mathbf{x}_n\}$ with hard labels $\mathbf{Y}=\{\mathbf{y}_1, \mathbf{y}_2, \ldots, \mathbf{y}_n\}$, label smoothing creates weak labels $\mathbf{\tilde{Y}}=\{\mathbf{\tilde{y}}_1, \mathbf{\tilde{y}}_2, \dots, \mathbf{\tilde{y}}_n\} $ by replacing hard labels (a single index $y_i \in [1, \ldots, \mathcal{C}]$ indicating a class) with softened labels $\mathbf{\tilde{y}}i = [\tilde{y}^1_i, \tilde{y}^2_i, \dots, \tilde{y}^\mathcal{C}_i]$, where $\sum_{c=1}^{\mathcal{C}} \tilde{y}^c_i = 1$, $0 \leq \tilde{y}^c_i \leq 1$ indicating degree of membership in classes. For instance, it's possible for a person who is walking slowly to have sensor readings that resemble those of someone who is standing still, which can result in mislabeling. In such cases, an instance may be labeled as [0.4, 0.6] instead of hard labeling of [0, 1], due to the ambiguity in the sensor data. These weak labels preserve semantic relationships while acknowledging ambiguity and uncertainty in the labels. 

To seed train a model, \method~ uses label smoothing via constructing a mixing matrix $\mathbf{M}$, which is a linear combination of the identity matrix $\mathbf{I}$ and an all-ones matrix $\mathbf{J}$, controlled by a parameter $\alpha$~\citep{lukasik2020does}. Specifically, $\mathbf{M} = (1 - \alpha) \cdot \mathbf{I} + \frac{\alpha}{N} \cdot \mathbf{J}$, where $N$ is the number of classes. The resulting matrix $\mathbf{M}$ is then used as the target distribution in the categorical cross-entropy loss function during training, instead of using the one-hot encoded labels. As weight averaging is one of the central components of~\method, we also leverage exponential moving average (EMA) as an additional regularizer~\citep{izmailov2018averaging}. EMA maintains a running average of the model weights calculated as a decaying average of previous weights and current iteration weights. This running average smooths out the rapid fluctuations in weights over the course of training; making the model less sensitive to the specific instances and providing robustness against overfitting.

\textbf{Refinement with Few-shot Label Acquisition.}
To initiate the fine-tuning phase of the seed model $f_{\mathcal{B}}(\cdot)$, we propose a human-in-the-loop mechanism to obtain expert (or clean) labels for a small number of instances, which is largely cost-effective since only a few examples have to be presented to a human expert for labeling.  Formally, given a dataset $D$ with instances $(\mathbf{x}_i, \mathbf{y}_i)$, we select a subset $\mathcal{S} \subset \mathcal{D}$ and obtain expert labels $\mathbf{\widehat{y}}_i$ for each instance $\mathbf{x}_i \in \mathcal{S}$, resulting in $\mathcal{D}_{\mathbf{e}}$. These labeled examples can then be used to adapt the existing pretrained model with EMA as earlier stage. We note that directly training on a smaller subset of examples result in a model of subpar quality. Formally, $f_{\mathcal{T}_{\theta}} = \text{Fine-tune}(f_{\mathcal{B}_{\theta}}, \mathcal{D}_{\mathbf{e}}, \eta)$, where $\eta$ represents the learning rate during fine-tuning (FT). The use of seed training with smoothed labels and transfer learning is critical to the success of this few-shot fine-tuning stage as it help reduce the amount of time and resources required to learn high-quality model from scratch and enable an effective way to leverage limited expert-labeled data. 

\textbf{Model Merging.} \label{subsubsec: mm}The last phase involves the merge of learned models from earlier stages, and as such, we propose a simple yet effective approach that merges the seed $f_{\mathcal{B}_{\theta}}$ and fine-tuned $f_{\mathcal{T}_{\theta}}$ models into an aggregate one with increased performance. The key idea is to perform a weighted average of the parameters of the constituent models to combine them into a unified model. This technique assumes that the models are structurally identical and share the initialization, which allows for the direct comparison and averaging of their parameters. 

Formally, let $\mathcal{F} = \{f_{\theta_1}, f_{\theta_2}, \ldots, f_{\theta_n}\}$ be a set of $n$ trained neural network models ($n = 2$ in our case) with parameters $\{\theta_1, \theta_2, \ldots, \theta_n\}$ and fixed weights $\{w_1, w_2, ..., w_n\}$ for each model. The parameters of the merged model $\theta_{\text{merge}}$ are calculated as the weighted summation of the parameters of the $n$ models: $\theta_{\text{merge}} = w_1 \times \theta_1 + w_2 \times \theta_2 + \ldots + w_n \times \theta_n$. We perform a simple weighted average based on its success in previous works ~\citep{wortsman2022model}. The merged model thereby encapsulates the collective strengths of all the constituent models $\mathcal{F}$, with their contributions determined by the fixed predefined weights. The weighting applied to each model's parameters also allows us to emphasize a certain model and subtly adjust the weaker models, yielding a well-balanced combined model that strikes a balance. Furthermore, averaging the parameters has several benefits over simply selecting one as it provides a simple and scalable approach to harnessing an ensemble of models that reduces variability, resulting in a merged model that is more robust and generalizes better. 

\begin{table*}[t]
\small
\centering
\caption{Performance evaluation of~\method~on different tasks against a range of baselines. We report accuracy averaged over three trials along with standard deviation. $n_l$ refers to the noise level, where $n_l = 0$ is clean labeled data (i.e., original labels provided in the corresponding dataset) and we use fixed sparsity rate of $n_s = 0.2$. CE, LS, PL, LC, CL, Bi-T, and FL refers to cross-entropy, label smoothing, poly loss, logit clip, confident learning, bi-tempered and focal loss, respectively.}
\def\arraystretch{1.2}%
\label{tab:main_results}
\begin{subtable}[t]{0.48\textwidth}
    \centering 
    \resizebox{\textwidth}{!}{%
    \begin{tabular}{lcccc}
        \hline
        \textbf{Method} & $n_l = 0.0$ & $n_l = 0.2$ & $n_l = 0.4$ & $n_l=0.6$ \\ \hline
        CE & 79.7 $\pm$ 0.4 & 78.9 $\pm$ 0.7 & 61.3 $\pm$ 2.9 & 30.0 $\pm$ 12.2 \\
        LS & 79.3 $\pm$ 0.7 & 80.0 $\pm$ 0.9 & 62.6 $\pm$ 2.3 & 29.6 $\pm$ 13.2 \\
        Mixup & 78.7 $\pm$ 0.9 & 79.1 $\pm$ 0.4 & 62.9 $\pm$ 3.6 & 29.3 $\pm$ 12.6 \\
        PL & 79.3 $\pm$ 0.2 & 78.6 $\pm$ 1.1 & 62.1 $\pm$ 3.0 & 30.0 $\pm$ 12.2 \\
        Bi-T & 76.4 $\pm$ 0.4 & 77.5 $\pm$ 0.8 & 62.9 $\pm$ 5.2 & 27.5 $\pm$ 13.2 \\
        LC & 78.7 $\pm$ 1.3 & 79.0 $\pm$ 1.0 & 52.3 $\pm$ 12.6 & 30.2 $\pm$ 12.7 \\
        CL & 80.0 $\pm$ 0.5 & 78.9 $\pm$ 2.1 & 62.7 $\pm$ 3.3 & 29.1 $\pm$ 11.6 \\
        FL & \multicolumn{1}{c}{78.9 $\pm$ 0.5} & \multicolumn{1}{c}{78.4 $\pm$ 0.8} & \multicolumn{1}{c}{64.7 $\pm$ 2.4} & \multicolumn{1}{c}{29.1 $\pm$ 13.6} \\ \midrule
        FHLR & \textbf{80.6 $\pm$ 0.3} & \textbf{80.0 $\pm$ 0.6} & \textbf{76.9 $\pm$ 1.1} & \textbf{74.1 $\pm$ 0.5} \\ \hline
    \end{tabular}%
    }
    \caption{\textbf{Sleep Scoring}}
\end{subtable}
\hfill
\begin{subtable}[t]{0.48\textwidth}
    \centering
    \resizebox{\textwidth}{!}{%
    \begin{tabular}{lcccc}
        \hline
        \textbf{Method} & $n_l = 0.0$ & $n_l = 0.2$ & $n_l = 0.4$ & $n_l=0.6$ \\ \hline
        CE & 89.2 $\pm$ 1.5 & 85.6 $\pm$ 2.8 & 70.3 $\pm$ 10.8 & 44.4 $\pm$ 5.1 \\
        LS & 89.0 $\pm$ 0.8 & 85.3 $\pm$ 1.4 & 67.6 $\pm$ 11.0 & 36.9 $\pm$ 5.5 \\
        Mixup & 87.7 $\pm$ 1.5 & 89.2 $\pm$ 1.6 & 65.1 $\pm$ 10.8 & 39.6 $\pm$ 1.1 \\
        PL & 86.4 $\pm$ 1.6 & 84.7 $\pm$ 2.0 & 66.4 $\pm$ 9.6 & 39.3 $\pm$ 4.9 \\
        Bi-T & 85.9 $\pm$ 1.0 & 84.2 $\pm$ 2.1 & 68.3 $\pm$ 13.7 & 38.9 $\pm$ 8.1 \\
        LC & 85.9 $\pm$ 4.7 & 84.7 $\pm$ 4.7 & 67.0 $\pm$ 14.9 & 36.0 $\pm$ 7.6 \\
        CL & 87.7 $\pm$ 1.1 & 87.2 $\pm$ 0.9 & 73.0 $\pm$ 9.7 & 43.3 $\pm$ 7.6 \\
        FL & \multicolumn{1}{c}{84.8 $\pm$ 0.6} & \multicolumn{1}{c}{80.5 $\pm$ 3.8} & \multicolumn{1}{c}{64.8 $\pm$ 10.5} & \multicolumn{1}{c}{40.1 $\pm$ 7.2} \\
        \midrule
        FHLR & \textbf{91.2 $\pm$ 0.5} & \textbf{89.2 $\pm$ 0.5} & \textbf{85.6 $\pm$ 0.3} & \textbf{85.5 $\pm$ 0.4} \\ \hline
    \end{tabular}%
    }
    \caption{\textbf{Activity Recognition}}
\end{subtable}
\vfill
\begin{subtable}[t]{0.48\textwidth}
    \centering
    \resizebox{\textwidth}{!}{%
    \begin{tabular}{lcccc}
        \hline
        \textbf{Method} & $n_l = 0.0$ & $n_l = 0.2$ & $n_l = 0.4$ & $n_l=0.6$ \\ \hline
        CE & 92.9 $\pm$ 1.5 & 91.0 $\pm$ 1.6 & 75.2 $\pm$ 2.3 & 25.5 $\pm$ 3.7 \\
        LS & 94.1 $\pm$ 0.0 & 87.8 $\pm$ 2.2 & 69.3 $\pm$ 7.6 & 29.2 $\pm$ 3.3 \\
        Mixup & 91.9 $\pm$ 1.5 & 86.0 $\pm$ 5.4 & 71.3 $\pm$ 5.3 & 25.9 $\pm$ 5.9 \\
        PL & 93.9 $\pm$ 0.2 & 82.0 $\pm$ 4.5 & 68.6 $\pm$ 7.9 & 24.8 $\pm$ 4.2 \\
        Bi-T & 90.5 $\pm$ 1.8 & 85.4 $\pm$ 4.8 & 71.1 $\pm$ 3.3 & 26.8 $\pm$ 5.8 \\
        LC & 90.9 $\pm$ 0.6 & 84.5 $\pm$ 5.7 & 76.5 $\pm$ 0.5 & 27.0 $\pm$ 2.8 \\
        CL & 94.0 $\pm$ 0.3 & 91.7 $\pm$ 1.7 & 80.0 $\pm$ 8.3 & 41.0 $\pm$ 11.6 \\
        FL & \multicolumn{1}{c}{91.9 $\pm$ 2.4} & \multicolumn{1}{c}{89.6 $\pm$ 0.7} & \multicolumn{1}{c}{70.3 $\pm$ 3.9} & \multicolumn{1}{c}{27.9 $\pm$ 2.6} \\
        \midrule 
        FHLR & \textbf{94.7 $\pm$ 0.1} & \textbf{92.7 $\pm$ 0.7} & \textbf{89.3 $\pm$ 1.1} & \textbf{83.0 $\pm$ 4.5} \\ \hline
    \end{tabular}%
    }
    \caption{\textbf{Cardiac Arrhythmia}}
\end{subtable}
\hfill
\begin{subtable}[t]{0.48\textwidth}
    \centering
    \resizebox{\textwidth}{!}{%
        \begin{tabular}{lcccc}
            \hline
            \textbf{Method} & $n_l = 0.0$ & $n_l = 0.2$ & $n_l = 0.4$ & $n_l=0.6$ \\ \hline
            CE & 84.4 $\pm$ 0.8 & 76.9 $\pm$ 2.6 & 65.2 $\pm$ 6.5 & 29.3 $\pm$ 15.3 \\
            LS & 84.3 $\pm$ 0.8 & 77.6 $\pm$ 1.5 & 66.1 $\pm$ 2.3 & 31.4 $\pm$ 10.6 \\
            Mixup & 83.0 $\pm$ 0.5 & 77.7 $\pm$ 2.3 & 68.7 $\pm$ 2.6 & 33.2 $\pm$ 17.4 \\
            PL & 83.7 $\pm$ 1.9 & 77.4 $\pm$ 1.3 & 63.5 $\pm$ 3.6 & 30.5 $\pm$ 12.0 \\
            Bi-T & 80.0 $\pm$ 1.0 & 75.4 $\pm$ 2.0 & 67.2 $\pm$ 1.6 & 33.0 $\pm$ 17.6 \\
            LC & 81.8 $\pm$ 1.1 & 75.3 $\pm$ 3.2 & 69.0 $\pm$ 0.5 & 31.0 $\pm$ 8.4 \\
            CL & 83.4 $\pm$ 0.3 & 78.8 $\pm$ 0.4 & 66.4 $\pm$ 4.1 & 30.5 $\pm$ 7.6 \\
            FL & \multicolumn{1}{l}{82.0 $\pm$ 0.9} & \multicolumn{1}{l}{73.4 $\pm$ 3.2} & \multicolumn{1}{l}{61.7 $\pm$ 5.3} & \multicolumn{1}{l}{30.7 $\pm$ 9.9} \\
            \midrule
            FHLR & \textbf{86.2 $\pm$ 0.1} & \textbf{81.5 $\pm$ 0.8} & \textbf{77.5 $\pm$ 1.0} & \textbf{72.6 $\pm$ 3.3} \\ \hline
        \end{tabular}%
    }
    \caption{\textbf{Artifact Detection}}
\end{subtable}

\end{table*}

\section{Experimental Setup}
We conduct an extensive evaluation across different noise profiles on several publicly available datasets to determine the feasibility of our \method~method in comparison with several baselines ranging from regularization techniques to cross-validation based methods.

\subsection{Data and Tasks}
We focus on a broad range of high-stakes tasks involving health signals collected from wearables, including IMUs, EEG, and ECG, across four widely used benchmark datasets. We provide the details of each of the considered datasets below. 

\textbf{Sleep Scoring~\textnormal{(SS)}.} For sleep stage scoring with EEG, we use Physionet Sleep-EDF dataset~\citep{kemp2000analysis} consisting of $61$ polysomnograms. The dataset includes $2$ whole-night sleep recordings of EEGs from FPz-Cz and Pz-Oz channels, EMG, EOG, and event markers from $20$ subjects. The signals are provided at a sampling rate of $100$Hz, and sleep experts annotated each $30$-second segment into eight classes. The classes include Wake (W), Rapid Eye Movement (REM), N1, N2, N3, N4, Movement and Unknown (not scored). We applied standard pre-processing to merge N3 and N4 stages into a single class following the American Academy of Sleep Medicine standards, and removed the unscored and movement segments. We utilize the Fpz-Cz channel from an initial study that explored the effect of age on sleep in healthy individuals to categorize sleep into $5$ classes, i.e., W, REM N1, N2, and N3~\citep{kemp2000analysis}.

\textbf{Activity Recognition~\textnormal{(AR)}.} We use the Heterogeneity Human Activity Recognition (HHAR) dataset~\citep{stisen2015smart} to recognize activities in daily living from IMU signals (i.e., accelerometer and gyroscope) collected from a smartphone. In total, nine participants performed $6$ activities (i.e., biking, sitting, standing, walking, stairs-up, and stairs-down) for five minutes to obtain balanced class distributions. We employ the IMU signals collected from smartphones in our experiments and segment them into fixed-size windows of $400$ samples with $50\%$ overlap and only apply standard mean normalization to the input.

\textbf{Cardiac Arrhythmia~\textnormal{(CA)}.} For the task of arrhythmia detection, we consider the Ningbo dataset~\citep{zheng2022large}. Since the original data contain multi-labels of cardiac arrhythmia, we use those samples which only have one positive class as we focus on multi-class classification tasks. To that end, we use instances of four classes that have more than three thousand instances, where the final $4$ classes include atrial flutter, normal sinus rhythm, sinus bradycardia, and sinus tachycardia. For data pre-processing, we window the ECG signals into $5$-second segments with no overlap and use signals sampled at $500$Hz with $12$ channels.

\textbf{Artifact Detection~\textnormal{(AD)}.} We use the TUH Artifact EEG dataset which is part of the TUH EEG Corpus~\citep{obeid2016temple} to perform
artifact recognition, e.g., eye movements, which can be useful to decide if the signals are noise-free for downstream applications. The dataset contains EEG signals recorded at $250$Hz and annotated clinically into $5$ artifact classes. 
The TUH EEG Corpus is the most extensive publicly available corpus comprising of thousands of subjects and session recordings following the international $10-20$ system. 
We use $23$ channels for the $\text{01-tcp-ar}$ EEG reference setup, as per the approach in~\citep{Zanga2019PyEEGLab}.
We use segments of length $512$ samples and pad with zeros where necessary.

\begin{table*}[t]
\centering
\footnotesize
\caption{Comparison of~$\method$~against prior techniques with asymmetric label noise for a fixed noise level $n_l = 0.4$.} %
\label{tab:asym_noise}
\begin{tabular}{@{}lcccc@{}}
\toprule
\textbf{Method} & \multicolumn{1}{l}{\textbf{Sleep Scoring}} & \multicolumn{1}{l}{\textbf{Activity Recognition}} & \multicolumn{1}{l}{\textbf{Cardiac Arrhythmia}} & \multicolumn{1}{l}{\textbf{Artifact Detection}} \\ \midrule
CE & 52.4 $\pm$ 1.7 & 62.0 $\pm$ 7.2 & 59.5 $\pm$ 8.4 & 57.0 $\pm$ 5.4 \\
LS & 54.8 $\pm$ 0.5 & 57.6 $\pm$ 2.7 & 67.8 $\pm$ 2.2 & 54.6 $\pm$ 4.9 \\
Mixup & 47.5 $\pm$ 11.0 & 51.3 $\pm$ 2.2 & 58.3 $\pm$ 3.3 & 57.3 $\pm$ 10.1 \\
PL & 44.8 $\pm$ 10.3 & 53.2 $\pm$ 3.7 & 52.1 $\pm$ 9.6 & 54.7 $\pm$ 8.0 \\
Bi-T & 40.5 $\pm$ 13.9 & 59.1 $\pm$ 2.7 & 59.1 $\pm$ 6.4 & 55.7 $\pm$ 7.1 \\
LC & 46.1 $\pm$ 3.3 & 53.8 $\pm$ 2.6 & 53.7 $\pm$ 9.3 & 62.2 $\pm$ 4.3 \\
CL & 52.0 $\pm$ 12.2 & 57.5 $\pm$ 3.6 & 70.8 $\pm$ 1.5 & 59.6 $\pm$ 7.1 \\
FL & 52.9 $\pm$ 4.7 & 56.6 $\pm$ 0.9 & 62.9 $\pm$ 7.8 & 59.0 $\pm$ 9.3 \\ \midrule
FHLR & \textbf{77.6 $\pm$ 0.6} & \textbf{82.2 $\pm$ 2.5} & \textbf{90.9 $\pm$ 0.9} & \textbf{78.6 $\pm$ 0.6} \\ \bottomrule
\end{tabular}%
\end{table*}

\subsection{Implementation Details}
We employ a $1$D convolutional architecture that operates on the temporal input signal. It consists of six repeated convolutional blocks with kernels of size $8$, $8$, $8$, $6$, $6$, and $4$, filters ranging from $24$, $32$, $64$. $72$, $96$ and $128$ in each convolutional layer, respectively. We use group normalization after the convolutional layers with $4$ groups except for the one after an input layer that has a group equal to the number of input channels.  We apply an ELU activation, max-pooling layers with pool size $8$, and a stride of $2$ is used after even blocks. To aggregate the features, we apply global average pooling of convolutional features and a final dense layer with the number of units matching the classes. We apply L2 regularization with a coefficient of $10^{-4}$ and a dropout layer after the last convolutional block with a rate of $0.15$ to prevent overfitting. 

We train the model with the Adam optimizer and learning rate of $0.001$ until convergence on the training set. To generate smooth labels for seed model training we use a smoothing factor of $\alpha=0.05$. For exponential moving average of model weights during both FHLR training phases, we use a momentum of $0.99$. In the refinement phase, we acquire labels for $100$ examples selected in a stratified manner from an oracle, unless mentioned otherwise. In the last model merging stage, we apply a weighted average of the seed and fine-tuned models with the weight value of $w_\mathcal{B} = 0.15$ for the seed model in high noise profile and $w_\mathcal{B} = 0.9$ low noise profiles. The selection of these values is based on the observation that in case of high noise it is feasible to stay close to fine-tuned model and vice versa and can be selected using a validation set. In future work, we aim to investigate a more principled approach to selecting $w$ that further improves generalization. 
 
\subsection{Baselines and Evaluation}
We compare our method with several baselines spanning loss correction, data augmentation, pruning noisy examples via cross-validation and more. Specifically, we include bi-tempered (Bi-T) loss~\citep{amid2019robust}, label smoothing (LS)~\citep{lukasik2020does}, mixup~\citep{zhangmixup}, poly loss (PL)~\citep{lengpolyloss}, confident learning (CL)~\citep{northcutt2021confident}, logit clip (LC)~\citep{wei2022logit}, focal loss (FL)~\citep{lin2017focal}, and standard cross-entropy (CE) objective. For performance evaluation, we divide the HHAR and SleepEDF datasets into train and test splits (a $70$:$30$ ratio) with disjoint user groups and no overlap. For Ningbo and TUH Artifact EEG datasets, we perform a standard random split. In all cases, we report accuracy averaged over three independent trials.

\section{Results}
\noindent \textbf{Comparative analysis with baseline methods.} We begin the evaluation of our method by comparing it with several baselines across four tasks in Table~\ref{tab:main_results}. We vary the noise levels from $0$ to $0.6$ while keeping sparsity fixed at $0.2$. We demonstrate that our method achieves strong performance across all tasks and noise levels compared to prior techniques. On the sleep scoring task,~\method~attains $80.6$\% accuracy with no noise, outperforming all baselines. More importantly, it maintains $74.1$\% accuracy with a noise level as high as $0.6$, substantially higher than other baselines, where LC achieves only $30.2$\%. Furthermore, we observe similar trends in activity recognition, cardiac arrhythmia, and artifact detection tasks. We note that other methods, including data augmentation and loss correction, suffer substantially as noise levels increase. On the other hand, \method~consistently outperforms baselines, with absolute improvements up to $43$\% compared to baselines at $0.6$ noise level for sleep scoring.

\begin{figure*}[t]
    \centering
    \includegraphics[width=0.8\textwidth]{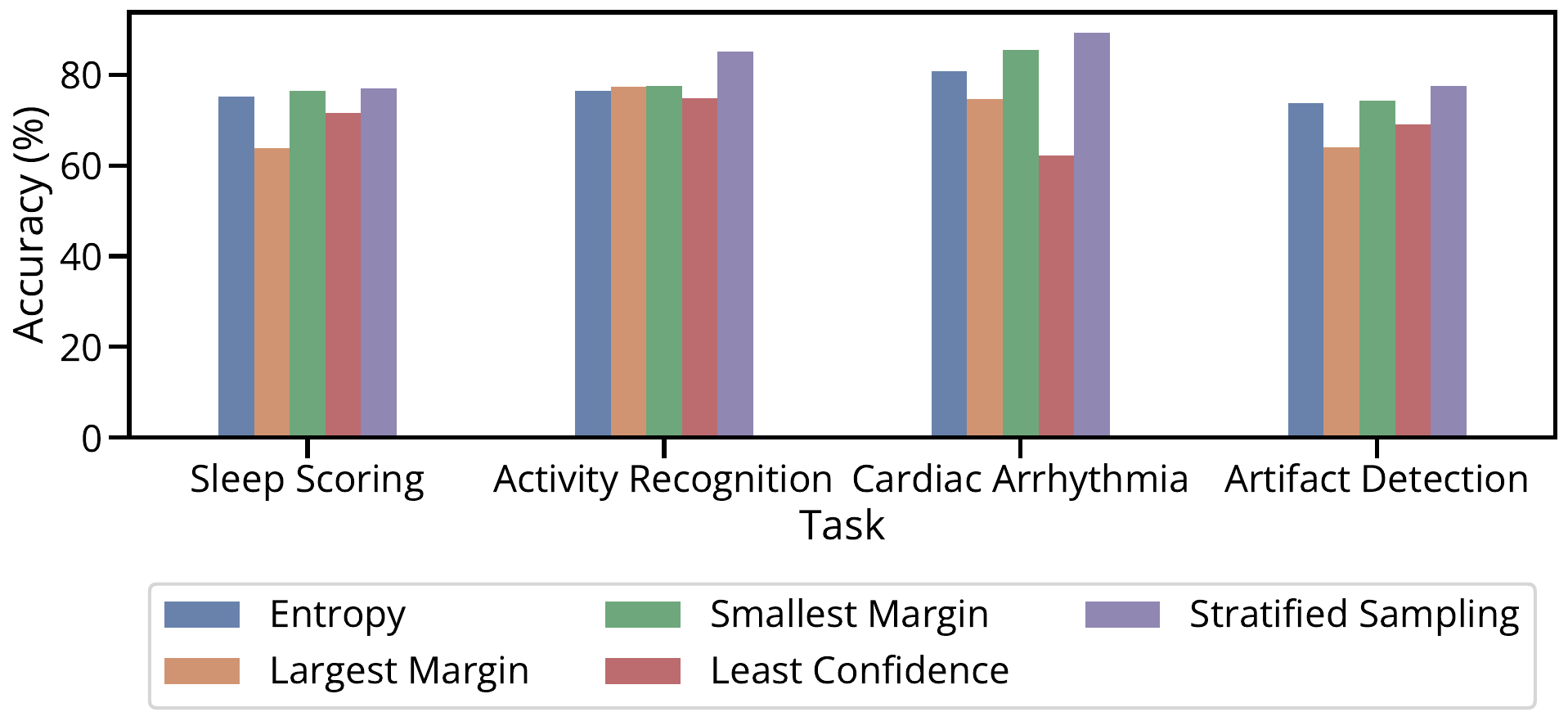}
    \caption{Ablation of different label acquisition strategies.}
    \label{fig:sampling_methods}
\end{figure*} %

\noindent \textbf{Robustness to asymmetric label corruption.} We next consider the evaluation of our approach on asymmetric label noise, i.e., when noise sparsity equals one. In this case, a special case of `class-flipping' occurs, where instances of a pair of classes are confused~\citep{northcutt2021confident}. The high sparsity noise could be attributed to confusion between classes that are perceived as similar by humans, e.g., mislabeling walking upstairs as walking rather compared to structurally different activities like sitting or cycling. 

Table~\ref{tab:asym_noise} summarizes the results across four tasks and seven baselines. Our method achieves the best performance in terms of classification accuracy, demonstrating its effectiveness in dealing with asymmetric noise. Compared to the cross-entropy objective, FHLR yields an improvement of $25$\% on sleep scoring, $20$\% on activity recognition, $31$\% on cardiac arrhythmia and $21$\% on artifact detection. On the other hand, other approaches show limited robustness against noise with CL achieving $70.8\%$ on cardiac arrhythmia while being several folds more computationally expensive as well as discarding valuable data. Similarly, label smoothing on its own is not sufficient, indicating that its effectiveness is limited to asymmetric label noise.

\noindent \textbf{Effect of expert label acquisition techniques.} The \textit{few-shot label acquisition for refinement} phase is a central component of our approach. To that end, we study the impact of different strategies for selecting samples for which we acquire labels from a human expert or an oracle. We perform evaluation with a noise level of $n_l = 0.4$ and sparsity of $n_s = 0.2$ and compare stratified (random) sampling with different uncertainty-based techniques, including entropy, smallest margin, largest margin, and least confidence. Figure~\ref{fig:sampling_methods} presents results across four tasks when only the label acquisition function is changed while keeping the rest of the components of~\method~fixed. We acquire labels for $100$ instances as in previous experiments. 

\begin{figure}[!t]
    \centering
    \includegraphics[width=0.8\textwidth]{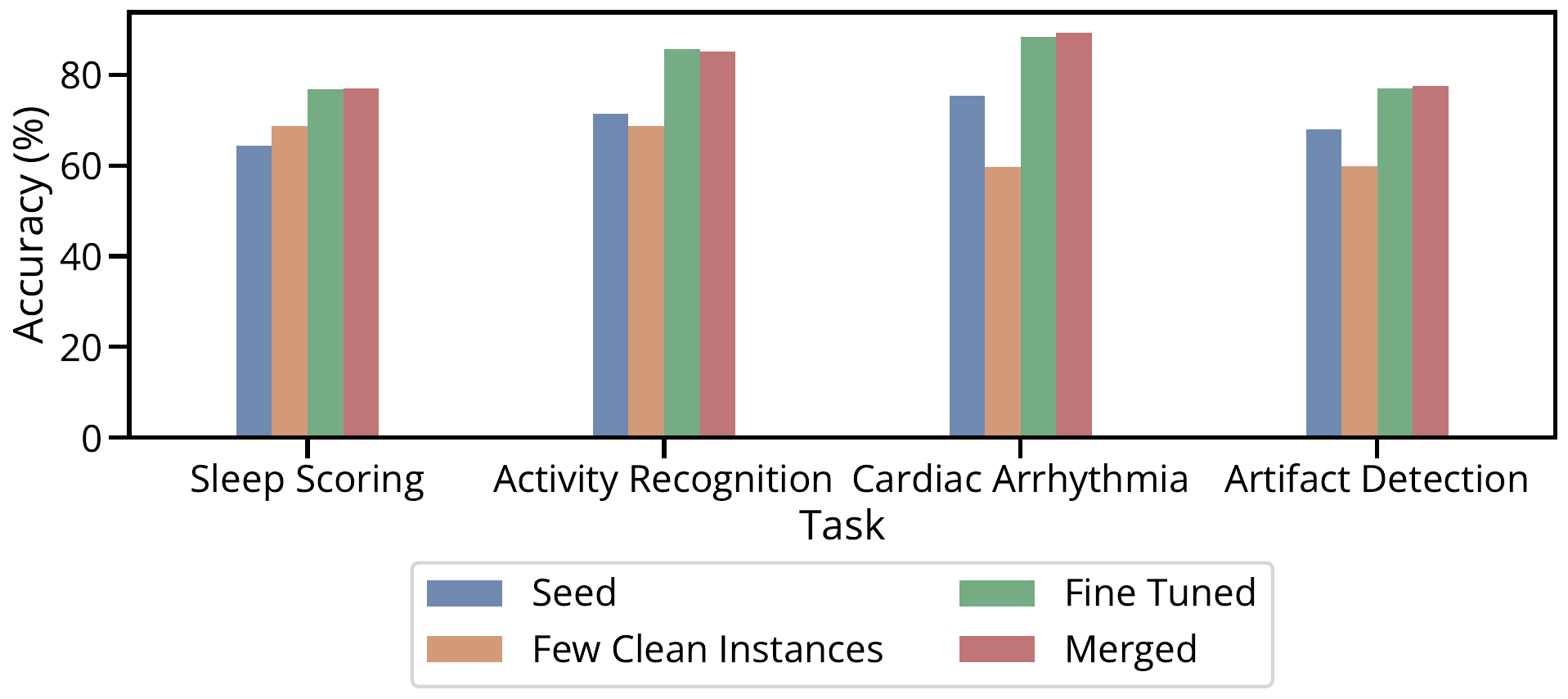}
    \caption{Performance improvement with model merging. }
    \label{fig:merged_vs_others}
\end{figure}

To make for a fair comparison with stratified sampling, we select instances based on classes using labels predicted by the seed model to avoid over-selection of instances from a particular class. We observe that entropy performs well compared to other approaches on all tasks except for activity recognition. In particular choosing examples based on least confident (i.e., lowest softmax probability) does not yield selection of quality instances. Our approach of randomly selecting examples in a class-balanced manner provides better performance across the board. 

\begin{table}[!t]
\centering
\caption{Comparison of model merging via parameters averaging against ensembles and using Fisher-weighted averaging~\citep{matena2022merging}.}
\label{tab:model_mergeing}
\def\arraystretch{1.2}%
\begin{tabular}{lllc}
\hline
\textbf{Task} & \multicolumn{1}{c}{\textbf{Ensemble}} & \multicolumn{1}{c}{\textbf{Fisher}} & \textbf{Conventional} \\ \hline
Sleep Scoring & 76.3 $\pm$ 2.1 & 76.5 $\pm$ 2.2 & 76.9 $\pm$ 1.1 \\
Activity Recognition & 83.3 $\pm$ 2.9 &  85.6 $\pm$ 0.6  & 85.1 $\pm$ 2.1 \\
Cardiac Arrhythmia & 90.2 $\pm$ 1.1 &  88.7 $\pm$ 0.5  & 89.3 $\pm$ 1.1 \\
Artifact Detection & 78.3 $\pm$ 1.0 &  76.7 $\pm$ 1.3  & 77.5 $\pm$ 1.0 \\ \hline
\end{tabular}%
\end{table}

\noindent \textbf{Effectiveness of model merging.} We now study the effect of model merging in improving generalization under label noise. Figure~\ref{fig:merged_vs_others} provides the results of the evaluation and compares merging with a seed model, fine-tuning with a few expert labels, and directly training a model from scratch using only a few labeled examples. Our results highlight that parameter averaging of seed and fine-tuned models improves performance; demonstrating that simple parameter averaging is useful in leveraging fine-tuned models and priors from the seed model to mitigate the effects of label noise.

\begin{figure*}[t]
    \centering
    \includegraphics[width=0.4\textwidth]{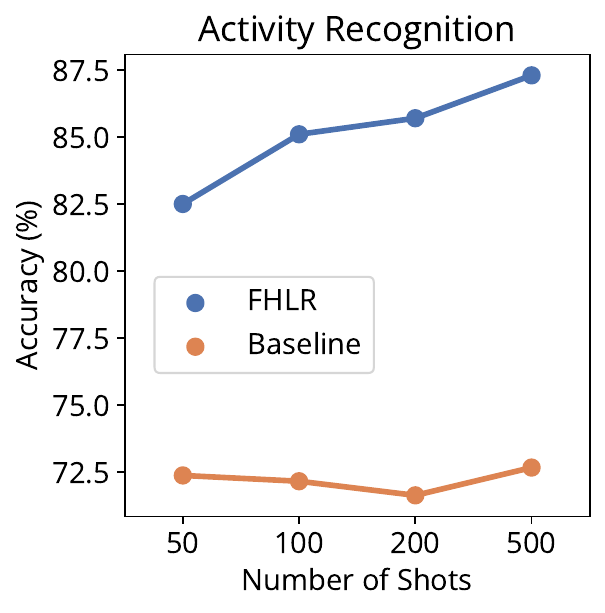} 
    \includegraphics[width=0.4\textwidth]{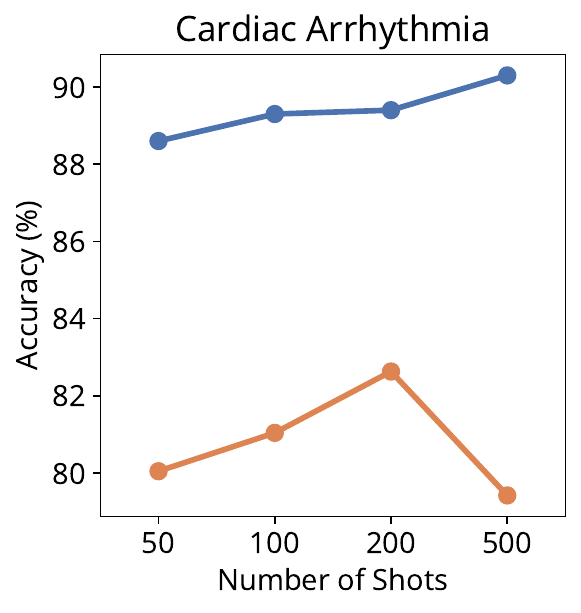}
    \caption{Ablation of a varying number of shots, i.e., few clean examples.}
    \label{fig:nb_clean_examples}
\end{figure*}

\noindent \textbf{Power of plain parameter averaging.} Table~\ref{tab:model_mergeing} reports the performance of different techniques to combine the models. On the considered tasks, conventional parameter averaging performs as well as model ensembles and Fisher-weighted averaging. For instance, on the sleep scoring task, the ensemble approach yields $76.3$\% accuracy, while Fisher-weighted averaging achieves $76.5$\%. In contrast, our approach, which simply averages model parameters, attains higher accuracy of $76.9$\%. Overall, these results suggest that `conventional' parameter averaging can be highly effective for merging deep models than more complex techniques. The conventional approach likely benefits from preserving more representative parameters through simple averaging without increasing inference costs as ensembles.

\noindent \textbf{Scaling up expert labels yields better generalization.} We next vary the number of clean labeled examples acquired during the refinement phase. For a noise level of $n_l=0.4$ and sparsity $n_s=0.2$, we conduct an experiment to get labels for randomly selected instances $\mathcal{S}$ as earlier while only changing the number of examples (or shots). We also compare against the best-performing baseline from Table~\ref{tab:main_results} (i.e., confident learning for activity recognition and cardiac arrhythmia) and correct the labels for the same number of instances. Figure~\ref{fig:nb_clean_examples} presents the evaluation results indicating 
that, while scaling the number of corrected labels consistently improves the performance of our method, the baseline method does not show a substantial and tangible performance boost when trained with an equal number of clean samples.

\noindent \textbf{Component-wise analysis validates utility of~\method.} We conduct an ablation to quantitatively demonstrates the utility of our three-stage method. Table~\ref{tab:component_ablation} provides the result on artifact detection task for a same configuration as used for the preceding experiment. Utilizing label smoothing alone results in $63.8$\% accuracy. The addition of exponential moving average provides further gains to $68.0$\%. Fine-tuning model parameters leads to a substantial $10$\% absolute improvement, achieving $74.5$\% accuracy. By incorporating proposed techniques model attains the highest accuracy of $77.5$\%, highlighting the resilience of~\method~in learning under label noise.
\begin{table}[!htbp]
  \centering
  \begin{tabular}{ccccc}
  \toprule
  \multicolumn{4}{c}{\textbf{Components}} &
    \multicolumn{1}{c}{\textbf{Accuracy}}       \\
    LS & EMA & FT & Merge &    \\
    \midrule 
     \checkmark &  & & &  63.8 $\pm$ 10.1      \\
     \checkmark & \checkmark &  & & 68.0	$\pm$ 6.7    \\
     \checkmark &  & \checkmark &  &  74.5 $\pm$ 1.7 	 \\
     \checkmark & \checkmark & \checkmark & &  76.9 $\pm$ 0.9 \\
     \checkmark &  & \checkmark & \checkmark & 75.5 $\pm$ 1.6    \\ \midrule
     \checkmark & \checkmark & \checkmark & \checkmark & 77.5 $\pm$ 1.0  \\ 
    \bottomrule
  \end{tabular}
   \caption{Ablating key components of~\method~on the artifact detection task.}
   \label{tab:component_ablation}
\end{table}

\noindent \textbf{Impact of annotators disagreement in refinement stage.} We further conduct an experiment to showcase the effectiveness of \method~in a more realistic scenario where there is no single source of ground truth. As is common in wearable datasets~\citep{sabeti2019signal}, multiple annotators have disagreements that can have an impact on the refinement phase. To study this, we introduced $10$ virtual annotators (in the refinement phase) by varying the disagreement rate using the Fleiss Kappa and compared it to our baselines (see Table~\ref{tab:main_results} where disagreement rate = $0$) for noise level of $0.4$ and sparsity of $0.2$ in Table~\ref{tab:virtual_annot}. Our results demonstrate the practical benefits of~\method~when learning from datasets that are not straightforward to annotate because they require domain expertise. The resilience of our method is evident in scenarios characterized by potential annotator disagreement. Even when confronted with such variability in the labeling of datasets for sleep scoring and activity recognition tasks, our approach successfully circumvents catastrophic failure, maintaining a high level of performance with a minimal decrease of only $2$\%. This indicates that our method ensure reliable model training even under less-than-ideal conditions.

\begin{table}[!t]
\centering
\small
\begin{tabular}{@{}lccc@{}}
\toprule
\textbf{Task} & \textbf{Disagreement Rate} & \textbf{Accuracy} & \textbf{Fleiss Kappa} \\ \midrule
\multirow{2}{*}{Sleep Scoring} & 0.1 & 76.0$\pm$1.6 & 75.04 \\
 & 0.2 & 75.8$\pm$1.6 & 53.84 \\
\hline
\multirow{2}{*}{Activity Recognition} & 0.1 & 82.9$\pm$1.9 & 77.57 \\
 & 0.2 & 80.1$\pm$2.4 & 57.69 \\
 \hline
\multirow{2}{*}{Cardiac Arrhythmia} & 0.1 & 88.1$\pm$1.4 & 74.07 \\
 & 0.2 & 86.7$\pm$1.5 & 52.49 \\
 \hline
\multirow{2}{*}{Artifact Detection} & 0.1 & 75.2$\pm$0.9 & 74.30 \\
 & 0.2 & 73.4$\pm$1.2 & 53.39 \\ \bottomrule
\end{tabular}
   \caption{Simulating human expert disagreement in the refinement phase.}
   \label{tab:virtual_annot}
\end{table}

\section{Conclusion}
This work proposed FHLR, a novel approach to mitigate the impact of label noise by learning from weak labels, incorporating human expertise and model merging. FHLR achieves significantly better generalization compared to several prior techniques across four tasks. It provides an effective way to overcome label noise without assumptions on noise distribution or extra components, using only a modest number of verified labels. This enables building robust models for health monitoring using wearables in the presence of annotation noise. Overall, our approach has important implications for advancing deep learning with noisy labels in various real-world applications.

\bibliography{main}
\bibliographystyle{plainnat}

\end{document}